\documentclass[11pt]{article}

\usepackage[margin=1in]{geometry}
\usepackage[T1]{fontenc}
\usepackage[utf8]{inputenc}
\usepackage{microtype}
\usepackage{booktabs}
\usepackage{amsmath}
\usepackage{caption}
\usepackage{titlesec}
\usepackage{xcolor}

\definecolor{myblue}{rgb}{0.0,0.25,0.55}

\usepackage[
    colorlinks=true,
    linkcolor=black,
    citecolor=black,
    urlcolor=myblue
]{hyperref}

\titlespacing*{\section}{0pt}{1.4ex plus 1ex minus .2ex}{0.8ex plus .2ex}
\titlespacing*{\subsection}{0pt}{1.1ex plus 1ex minus .2ex}{0.6ex plus .2ex}

\newcommand{\pct}{\%}

\title{\textbf{TamilEOT: A Dataset and Model for Semantic\\End-of-Turn Detection in Tamil Telephone Speech}}

\author{Santhoshkumar V\\[2pt]
\small \href{mailto:santhoshv.work@gmail.com}{santhoshv.work@gmail.com}\\[2pt]
\small \url{https://github.com/santhosh-005/tamil-eot}}

\date{}

\begin{document}
\maketitle

\begin{abstract}
\noindent
A voice agent must decide, at each pause, whether the speaker has finished or
is still thinking. Without a model of the language, systems fall back to a
fixed silence timeout: too short, and the agent interrupts; too long, and every
turn pays the full wait. We release TamilEOT, to our knowledge the first open
dataset and detector for semantic end-of-turn detection in a South Indian
language: 18,485 labelled turn boundaries from 116 real Tamil telephone
conversations, two audio-only detectors built on Smart Turn v3's architecture,
and a turn-level companion set of 4,774 whole turns with their pause structure.
On a held-out split of 4,168 clips from 30 unseen calls, accuracy rises from
70.30\pct{} zero-shot to 83.71\pct{} (8.7\,MB) and 86.13\pct{} (21\,MB); ROC-AUC
rises 0.751 $\rightarrow$ 0.921. Both run in under 150\,ms single-threaded on a
laptop CPU.

We also report what building it cost. Our rule-derived negative-class labels
scored 44.4\pct{} against a blind human pass --- below chance --- because the
rule answered a different question than the model is asked; an audio-LLM
labeller at 97.5\pct{} agreement replaced them for US\$5.69. Of six training
levers, only encoder capacity moved the result: three identical runs spanned
0.87 accuracy points, the floor below which nothing else here is a finding.
Replaying labelled boundaries through the production VAD and streaming adapter
costs a further 2.60 points, and 7.8\pct{} of boundaries never reach the model
at all. Data, weights, code and every negative result are public.
\end{abstract}

\section{Introduction}
\label{sec:intro}

At each pause, a voice agent must decide whether the speaker has finished or is
still thinking. Voice activity detection cannot make that call: it detects
sound, not intent. Systems built on VAD alone fall back to a fixed silence
timeout, a single dial with two bad ends: too short, and the agent talks over a
speaker who paused mid-sentence; too long, and every completed turn waits for a
timer that exists only to catch the pauses.

A semantic end-of-turn (EOT) detector removes the dial. It listens to the last
few seconds of the speaker's audio and predicts completion from prosody, without
waiting for a transcript. Smart Turn v3 \cite{smartturn} is the most widely
deployed open model of this kind, and its released weights cover a fixed
language list that does not include Tamil. This paper is about detectors that
can be inspected, retrained and run locally, where a fixed language list is a
hard limit rather than a feature request.

This paper describes what it took to close that gap for one language, and what
we got wrong on the way.

\paragraph{Contributions.}
\begin{enumerate}\itemsep2pt
\item \textbf{A dataset.} 18,485 labelled turn boundaries from 116 real Tamil
narrowband telephone conversations, split by call, released CC BY 4.0
(Section~\ref{sec:data}). To our knowledge it is the first open dataset for
semantic end-of-turn detection in any South Indian language. From the same
calls we also release 4,774 whole turns with their
internal pause structure, for the turn-taking questions an 8-second window
cannot express (Section~\ref{sec:turns}).

\item \textbf{A validated labelling protocol.}
We measured our own rule-derived labels against a blind human pass, found the
negative class below chance, diagnosed why, and replaced it with an audio-LLM
labeller chosen by measurement. We also report the resulting label-noise
ceiling (Section~\ref{sec:labels}).

\item \textbf{Two models and a set of negative results.} 70.30\pct{}
$\rightarrow$ 86.13\pct{} on the held-out split. Of six training levers, one
worked. We publish the five that did not, together with the run-to-run spread
that makes small claims unfalsifiable (Section~\ref{sec:results}).

\item \textbf{A measurement of what serving costs.} Offline accuracy scores a
clip cut by the dataset builder; production scores a window cut by a VAD. We
replayed the same labelled boundaries through the real VAD and the real
streaming adapter and measured the difference (Section~\ref{sec:serving}).
\end{enumerate}

\paragraph{A note on conventions.}
Two false-positive rates are in circulation and they differ by roughly
$3\times$. Smart Turn publishes $\mathrm{FP}/N$, where $N$ is the whole
evaluation set, so that $\mathrm{FP}/N$ and $\mathrm{FN}/N$ sum to the error
rate. The standard $\mathrm{FPR}=\mathrm{FP}/(\mathrm{FP}+\mathrm{TN})$ is
computed over negatives only. On our test set 10.1\pct{} in the first convention
is 27.4\pct{} in the second. Every table below states which one it uses. The
positive class throughout is \texttt{complete}, so a false positive is
\emph{the agent talking over the user} --- the expensive error.

\section{Related work}

\paragraph{Turn-taking models.} Voice Activity Projection
\cite{vap} learns turn-taking events self-supervised from the future voice
activity of both speakers, and needs both channels. Smart Turn v3
\cite{smartturn} takes the deployment-shaped version of the problem: one
channel, the user's, and a binary verdict on the last 8 seconds. It is a
Whisper-tiny encoder \cite{whisper} with a classification head, about 8M
parameters, quantised to int8 and running in tens of milliseconds on CPU. Weights,
data and training code are all open, which is why we replicated it rather than
designing our own. Its released v3 weights cover 23 languages; the three Indic ones
(Bengali, Hindi, Marathi) are all Indo-Aryan, and no South Indian language is
included.

\paragraph{Datasets.} The ETD dataset \cite{etd} is, to our knowledge, the first
public corpus aimed directly at end-turn detection, combining TTS-generated
dialogue with real conversational audio. Smart Turn's own training corpus is largely TTS. Neither
covers Tamil, and neither is narrowband telephone speech, which is the acoustic
condition most Indic voice agents actually run in.

\paragraph{Language-specific work.} The closest analogue to this paper is a
recent study of Thai end-of-turn detection \cite{thaieot}, which builds a Thai
baseline from transcribed subtitles and classifies at token boundaries. That
work is text-only and depends on a transcript; ours is audio-only and runs
before one exists. The two approaches are complementary: a transcript-based
detector cannot fire until ASR has emitted, and an audio-based one cannot read
lexical cues.

\section{Dataset}
\label{sec:data}

\subsection{Source, and why this corpus}

The audio is the \texttt{ta\_IN\_*\_\{Left,Right\}} family of
SPRING-INX Tamil R1 \cite{springinx}, released CC BY 4.0 by SPRING Lab, IIT
Madras. It is 116 two-party Tamil telephone conversations, and it has one
property that decided the whole project: \textbf{each conversation ships as one
audio file per speaker}, each leg transcribed separately by a human.

Which file the audio came from \emph{is} the speaker label. There is no
diarization step and therefore no speaker-error rate propagating into the
labels. We verified 116/116 pairs complete with 0\,ms duration mismatch between
legs. Every clip is cut from one leg only, which is also what a deployed
detector receives: a voice agent sees the inbound user stream, not a mixdown.

\subsection{Boundaries come from the VAD, not the transcript}

The obvious way to find turn boundaries is to use the transcript segment edges.
That does not work here, because the shipped segments are padded. Measuring the
same quantity two ways shows it directly: cross-channel overlap is 28.1\pct{} of
speech when computed from segment boundaries, but only 9.4\pct{} when computed
from VAD speech spans on the same audio. A segment that begins before its speaker
starts and ends after they stop manufactures overlap that is not there. Their
edges therefore inflate apparent overlap and misplace every gap.

We run Silero VAD \cite{silero} on each leg independently and take speech
onsets and offsets from it. The transcript is used for exactly two things:
confirming that a VAD span is real speech on \emph{that} leg (a span with no
transcript over it is crosstalk bleed from the other leg), and supplying text
for the backchannel filter. Words are apportioned across VAD speech seconds
rather than elapsed time --- the median segment is 21\pct{} silence and the
90th percentile is 53\pct{}, so linear interpolation would hand about a fifth
of the words to intervals where nobody spoke.

From 116 calls this yields 50,532 candidate boundaries: 20,013 floor changes
and 30,519 same-speaker holds.

\subsection{Clip geometry}

Each sample is the 8 seconds of that speaker's audio ending 200\,ms after the
speech offset. \textbf{Both classes get exactly the same 200\,ms of trailing
audio.} This matters more than it looks: if positives ended with more silence
than negatives, a model could learn to read silence length instead of speech,
score well offline, and collapse against a production endpointer with different
timing.

We test for it directly. \texttt{09\_verify.py} fits a
deliberately weak classifier on ten crude global features (energy, duration,
spectral tilt, voiced fraction) that carry no prosodic structure. Near-chance
performance is the pass condition. Results are in Section~\ref{sec:shortcut}.
On the most recent harvest, measured peak amplitude in the final 200\,ms is
median 0.0021 and p95 0.0225: the trailing window is silence, and its length
carries no label information.

\subsection{Composition and splits}

Boundaries fall into four provenance classes, by how they were detected and
whether the two available witnesses (the other speaker's behaviour and the
transcriber's segmentation) agreed (Table~\ref{tab:sources}). Provenance is
recorded on every row but is \emph{not} a feature: it is unavailable at
inference.

Two of the four classes (\texttt{change\_midseg}, \texttt{hold\_inter}) were
initially held back precisely because the witnesses disagreed and a rule had to
break the tie alone. Once a validated third witness existed
(Section~\ref{sec:labels}) they were released into the set. A second pass over
rejected boundaries with two gates relaxed added a further 2,272 rows.

\begin{table}[t]
\centering
\small
\begin{minipage}[t]{0.52\textwidth}
\centering
\begin{tabular}{llr}
\toprule
provenance & detected as & $n$ \\
\midrule
\texttt{hold\_intra} & negative & 10,402 \\
\texttt{change} & positive & 4,660 \\
\texttt{hold\_inter} & negative & 2,343 \\
\texttt{change\_midseg} & positive & 1,080 \\
\midrule
total & & 18,485 \\
\bottomrule
\end{tabular}
\caption{Provenance of the 18,485 boundaries. The shipped label is the
labeller's verdict, which overrules provenance on 53\pct{} of
\texttt{hold\_intra}.}
\label{tab:sources}
\end{minipage}\hfill
\begin{minipage}[t]{0.44\textwidth}
\centering
\begin{tabular}{lrrrr}
\toprule
split & calls & clips & compl. & incompl. \\
\midrule
train & 71 & 11,992 & 7,908 & 4,084 \\
dev & 15 & 2,325 & 1,532 & 793 \\
test & 30 & \textbf{4,168} & 2,629 & 1,539 \\
\midrule
total & 116 & 18,485 & 12,069 & 6,416 \\
\bottomrule
\end{tabular}
\caption{Splits, assigned \emph{by call}. The test split was cut once, before
any training, and was not revised afterwards.}
\label{tab:splits}
\end{minipage}
\end{table}

\paragraph{Splitting at the call level.} Splits are assigned per call rather
than per clip (Table~\ref{tab:splits}), so no \emph{call} in the test set was
seen during training. Both SPRING-INX releases ship utterance-level splits
instead, which place the same recording on both sides of a split; in the R2
release, 511 of 545 recordings appear in train and eval alike. A turn detector
trained across such a split can learn to recognise a recording in place of
prosody.

Call-disjoint is not speaker-disjoint, and we cannot claim the stronger property.
SPRING-INX ships no speaker identity --- \texttt{utt2spk} is degenerate, with
\texttt{speaker\_id == utt\_id} on all 94,281 rows --- so there is nothing to
verify it against. The 116 calls were recorded on seven days, one of which
contributes 38 calls, and calls from every one of those days appear in more than
one split. These are call-centre conversations in which one leg is an agent, and
agent-side clips are roughly half the data, so the same agent voices are likely
present on both sides of the split. We report this as a limitation rather than
estimating its effect, which would require speaker labels the corpus does not
provide.

\subsection{A turn-level companion release}
\label{sec:turns}

The 8-second window is the right unit for training an end-of-turn classifier and
the wrong one for almost everything else about turn-taking. It cannot represent a
speaker's own mid-turn pauses, the length of the silence that actually ends a
turn, or what the other speaker was doing across it. Those are the quantities a
timing or interruption study needs, and they are present in the same calls, so we
release them.

\texttt{tamil-turns} regroups the same audio into \textbf{4,774 whole turns} from
115 calls, 15.4 hours, carrying \textbf{9,709 mid-turn pauses}. Each row is one
speaker's continuous hold of the floor with every silence inside it recorded: the
last is the end of the turn, the earlier ones are hesitation. Turns come from
floor occupancy over the same VAD spans, not from a classifier, and six gates
decide which are clean: the turn opens a transcript segment; it closes one; the
other speaker does not talk across it; the handover is not overlapped; it lasts
at least a second; and it runs at no less than a word per second. Of
21,596 raw turns, 4,774 survive. The gate is deliberately strict: spot-listening the ungated
set turned up fragments cut mid-word and ``pauses'' that were the other speaker
holding the floor.

The split is the same call-level split used here, so the two releases agree on
which calls are held out. Unlike the clip set, this one is \emph{structural
rather than labelled}: the silence layout is derived from audio and transcript,
not from anyone's judgement, so no annotation was required. Where a labelled
boundary from the clip set lands on a turn's end the label is joined in, which
covers 2,113 of the 4,774 turns; the rest carry \texttt{null} rather than a
guess. Two properties limit what the release can be used for: turns the other
speaker cut into are excluded by the clean gate, so the set under-represents
interruption at the handover itself; and the per-turn text is apportioned across
VAD speech seconds rather than force-aligned, so it is approximate and carries no
word-level timings.

\section{Labels}
\label{sec:labels}

\subsection{Validating the rule-derived labels}

The first label set was rule-derived: a floor change at a segment end is
\texttt{complete}; a pause wholly inside one transcript segment is
\texttt{incomplete}. This is the standard construction, and it is cheap enough
that it is rarely questioned.

We ran a blind listening pass: 197 clips stratified by (provenance, label), no
label shown, no metadata, audio only, one listen each.

\begin{center}
\small
\begin{tabular}{llr}
\toprule
class & grounded in & agreement with listener \\
\midrule
\texttt{change} (positive) & the other person took the floor & \textbf{95.9\pct} \\
\texttt{hold\_intra} (negative) & a pause fell inside a segment & \textbf{44.4\pct} \\
overall & & 70.1\pct \\
\bottomrule
\end{tabular}
\end{center}

\textbf{44.4\pct{} is below chance.} The negative class was worse than a coin
flip.

\subsection{The rule answered a different question}

The positive class works because it is grounded in behaviour that was
\emph{recorded in the call}. The other speaker heard the turn end and took the
floor. That is a witness.

The negative class had no witness. ``A pause fell inside a transcript segment''
is not a judgement about completeness. It is a judgement about where a human
transcriber pressed enter.

Underneath this sits a genuine ambiguity, and naming it is the useful part.
Two defensible questions diverge on a speaker who finishes one sentence and
starts another:

\begin{center}
\small
\begin{tabular}{lll}
\toprule
& question & who asks it \\
\midrule
Q1 & is this utterance \emph{syntactically finished}? & our listeners, our prompt, Smart Turn's data \\
Q2 & does the speaker \emph{intend to continue}? & a pause oracle; transcript-derived labels \\
\bottomrule
\end{tabular}
\end{center}

Our rule-derived negatives answered Q2. Pause-derived labelling of real
conversation is a reasonable construction and prior work uses it
\cite{etd}; it simply is not the question a Smart Turn model is being
trained on. Rewording our evaluation to Q2 was ruled out for the same reason:
mixing a Q2 corpus into a Q1-pretrained model teaches two different questions.

We also confirmed that no amount of feature engineering rescues it. Every
structural, textual and metadata feature the pipeline computes, combined,
predicts the human verdict at AUC 0.637.

\subsection{Relabelling with an audio LLM}

The replacement had to \emph{listen}. We evaluated seven audio-capable LLMs on
the same 197 human-labelled clips, with the same prompt, audio only, no
transcript and no prior label, run paired through a batch API
(Table~\ref{tab:bakeoff}).

\begin{table}[t]
\centering\small
\begin{tabular}{lrrr}
\toprule
labeller & agreement & on \texttt{hold\_intra} & est.\ cost, full pool \\
\midrule
\textbf{\texttt{gemini-3.7-flash}} & \textbf{97.5\pct} & \textbf{97.0\pct} & \textbf{\$5.77} \\
\texttt{gemini-3.1-pro-preview} & 96.4\pct & 93.9\pct & \$41.17 \\
\texttt{gemini-3.6-flash} & 96.4\pct & 97.0\pct & \$9.10 \\
\texttt{gemini-3.5-flash} & 93.9\pct & 92.9\pct & \$11.82 \\
\texttt{gemini-3-flash-preview} & 93.4\pct & 90.9\pct & \$17.09 \\
\texttt{gemini-2.5-flash} & 91.8\pct & 90.8\pct & \$13.50 \\
\texttt{gemini-3.5-flash-lite} & 66.5\pct & 68.7\pct & \$1.18 \\
\midrule
\emph{the rule-derived labels} & 70.1\pct & 44.4\pct & --- \\
\bottomrule
\end{tabular}
\caption{Labeller bake-off on the 197 blind-listened clips. Six of seven clear
the 90\pct{} bar. By McNemar the top three are statistically indistinguishable
($p=0.75,\ 0.73$), so the tie was broken on price, not on the point estimate.}
\label{tab:bakeoff}
\end{table}

\texttt{gemini-3.7-flash} agrees with the human listener 97.5\pct{} of the time,
and 97.0\pct{} on the exact class the rules got wrong. It relabelled the full
pool for \textbf{US\$5.69} (a further US\$0.78 for the 2,272 rows added later).

Two independent measurements of the same error agree: \textbf{53\pct{} of the
negative class flipped} under relabelling, against \textbf{56\pct{}} in the
human pass. Class balance on the original pool moved from 4,751/11,462 to
10,493/5,720 complete/incomplete; the released set is 12,069/6,416.

Every row retains \texttt{label\_pipeline}, \texttt{llm\_verdict} and a
\texttt{dispute} flag, so the set can be re-derived under a different policy
without re-running anything.

\subsection{The confound this removed}
\label{sec:shortcut}

Relabelling also closed the one structural defect we knew about.
\texttt{hold\_intra} required a pause \emph{inside} a segment, so the negative
class was drawn from longer utterances by construction --- and that leaked into
the audio as voiced fraction. The gate was deciding the label, so the gate's
bias rode along with it.

\begin{center}\small
\begin{tabular}{lrrr}
\toprule
labels & shortcut-probe AUC & \texttt{prev\_dur} $d$ & \texttt{voiced\_frac} $d$ \\
\midrule
rule-derived & 0.601 & $-0.33$ & $-0.46$ \\
relabelled & 0.622 & $\mathbf{-0.09}$ & $\mathbf{-0.17}$ \\
\bottomrule
\end{tabular}
\end{center}

Both rows are the matched core subsets, so the comparison is like for like; on
the full released set \texttt{prev\_dur} $d=-0.11$. Cohen's $d$ on both leaked
features drops to near zero. The probe's AUC moves
\emph{up} slightly, 0.601 $\rightarrow$ 0.622; both figures sit inside the
near-chance band, and the class balance moved at the same time, so AUC is the
comparable quantity and accuracy is not.

\subsection{The label-noise ceiling}

Because both witnesses are recorded per row, the residual label noise can be
estimated directly from the QA sample.

\begin{center}\small
\begin{tabular}{lrr}
\toprule
subset & share of test & label accuracy \\
\midrule
both witnesses agree & 62.3\pct & 99.3\pct \\
disputed & 37.7\pct & 93.5\pct \\
\midrule
weighted & & \textbf{97.1\pct} \\
\bottomrule
\end{tabular}
\end{center}

\textbf{Measured accuracy much above 97.1\pct{} would mean agreeing with labeller
error.} The figure is an estimate, extrapolated from 197 clips scored by one
listener, and should be read with that provenance rather than as a hard ceiling.
Our best model is at 86.13\pct{}, so roughly 11 points of headroom remain on any
reading, and label quality is not what currently limits it.

Finally, the 197 human labels are themselves one listen to one isolated 8-second
clip with no future audio, and are not truth. Three were wrong: all seven
labellers contradicted them in the same direction, and on re-listening the
labellers were right. Corrected verdicts live in a file that every downstream
number is re-derived from, rather than being hardcoded.

\section{Model and training}
\label{sec:model}

The architecture is unchanged from Smart Turn v3: a Whisper encoder
\cite{whisper} over an 8-second log-mel window, attention pooling, and a binary
classification head. We train the whole stack, with no layers frozen.

\paragraph{Where the weights start.} We replicate Smart Turn v3's model class and
initialise the encoder exactly as its own training script
does --- from \texttt{openai/whisper-tiny} or \texttt{openai/whisper-base}, with
the pooling and classification heads randomly initialised. We do not fine-tune
from the released v3 checkpoint. Two reasons: upstream's \texttt{train.py} sets
\texttt{base\_model\_name} to \texttt{openai/whisper-tiny} and does the same, so
this is the recipe rather than a departure from it; and v3 is distributed as ONNX
only, with no trainable checkpoint published. Our models and Smart Turn v3 are
therefore siblings from a common initialisation, differing in training data, which
makes the zero-shot comparison in Section~\ref{sec:results} a comparison between
two independently trained models rather than a before-and-after on one.

Initialising instead from the released v3 weights, which would require recovering
them from the distributed ONNX graph, is a lever we did not take and do not
evaluate here.

\begin{center}\small
\begin{tabular}{ll}
\toprule
encoder & \texttt{openai/whisper-tiny} (8.0M) or \texttt{openai/whisper-base} (20.3M) \\
head & attention pooling $\rightarrow$ MLP binary classifier \\
loss & \texttt{BCEWithLogitsLoss} with per-batch \texttt{pos\_weight} \\
schedule & 6 epochs, lr $5\times10^{-5}$, batch 32, seed 0 \\
selection & best epoch on dev accuracy (epoch 3 or 4 in every run) \\
export & \texttt{.pt} $\rightarrow$ fp32 ONNX $\rightarrow$ int8 dynamic, on CPU \\
hardware & one NVIDIA T4 on Google Colab \\
\bottomrule
\end{tabular}
\end{center}

There is no multi-GPU step anywhere in this work. Both released models come from
one notebook and the encoder string is the only difference between them.

\paragraph{Are the released files the models we trained?} These are two
different claims and we checked both. First, both released ONNX files were
re-scored locally on the sealed test set and reproduce their training numbers.
Second, re-exporting the released \texttt{base} checkpoint produces a graph whose
predictions on the sealed test set are \textbf{numerically identical} to the
published one's ($\max|\Delta| = 0.00\mathrm{e}{+}00$; we compare outputs, not
serialised bytes).
Every number in this paper therefore comes from the artefact a reader can
download, not from a training log.

\section{Results}
\label{sec:results}

\subsection{The reproducibility floor}

Before any delta: three \texttt{base} runs at identical configuration and
identical seed scored \textbf{86.23\pct{} / 85.63\pct{} / 85.36\pct{}} --- a
spread of \textbf{0.87 points}.

\texttt{cudnn.deterministic} constrains only cuDNN ops. The encoder is
attention, routed through SDPA, whose flash and memory-efficient backward
kernels use atomics; non-associative float addition then produces divergent
trajectories over $\sim$2{,}244 optimiser steps. Checkpoint selection amplifies
it, since different runs peak at different epochs of a noisy curve.

\textbf{We therefore treat anything below about 1 point as noise}, and we ask
readers to do the same with every other number here. It is the single most
important figure in this paper for reading the rest of it.

\subsection{Main result}

\begin{table}[t]
\centering\small
\begin{tabular}{llrrrrrr}
\toprule
model & build & params & size & accuracy & ROC-AUC & FP/$N$ & p50, 1 thread \\
\midrule
majority class (\texttt{complete}) & --- & --- & --- & 63.08\pct & 0.500 & --- & --- \\
\texttt{smart-turn-v3.0}, zero-shot & int8 & 8.0M & --- & 65.64\pct & 0.779 & 7.08\pct & --- \\
\texttt{smart-turn-v3.2}, zero-shot & int8 & 8.0M & 8.7\,MB & 70.30\pct & 0.751 & 13.44\pct & --- \\
\midrule
TamilEOT-tiny & fp32 & 8.0M & 32\,MB & 83.35\pct & 0.904 & 7.73\pct & 133\,ms \\
\textbf{TamilEOT-tiny} & \textbf{int8 dyn} & 8.0M & \textbf{8.7\,MB} & \textbf{83.71\pct} & \textbf{0.905} & 7.94\pct & \textbf{83\,ms} \\
TamilEOT-tiny & int8 static & 8.0M & 8.4\,MB & 79.32\pct & 0.899 & 4.94\pct & 73\,ms \\
TamilEOT-base & fp32 & 20.3M & 81\,MB & 86.23\pct & 0.922 & 8.95\pct & 232\,ms \\
\textbf{TamilEOT-base} & \textbf{int8 dyn} & 20.3M & 21\,MB & \textbf{86.13\pct} & \textbf{0.921} & 9.17\pct & 143\,ms \\
TamilEOT-base & int8 static & 20.3M & 21\,MB & 73.46\pct & 0.792 & 13.70\pct & 125\,ms \\
\bottomrule
\end{tabular}
\caption{Test split: 4,168 clips from 30 held-out calls, threshold 0.5. FP/$N$
is Smart Turn's convention (see Section~\ref{sec:intro}). Latency is p50, batch 1, one thread,
inference only, measured on a single idle development laptop (Intel i5-12450H);
add ${\sim}12$\,ms for the mel front-end. Every build was timed the same way on
the same machine, so the column is comparable within itself; it is not a
deployment specification (Section~\ref{sec:serving}). Released models in bold.}
\label{tab:main}
\end{table}

Table~\ref{tab:main} is the headline. Training on Tamil takes accuracy from
70.30\pct{} to \textbf{86.13\pct}, a gain of \textbf{15.83 points}, and ROC-AUC
from 0.751 to 0.921.

The zero-shot row is worth reading closely, because it is what made the project
look tractable before anything was trained. At threshold 0.5, \texttt{v3.2}
calls \textbf{36.4\pct{} of unfinished Tamil turns finished}, so an agent using
it talks over the user on a third of their mid-sentence pauses. Its ROC-AUC,
however, is 0.751 rather than 0.5: Whisper's encoder already hears Tamil
prosody, the \emph{ranking} carries real signal, and only the decision boundary
is wrong. That is also why the two zero-shot rows disagree in the direction they
do. \texttt{v3.0} is the more conservative of the two, with a better AUC (0.779)
and roughly half the false completions (7.08\pct{} against 13.44\pct{}), but it
pays for that caution with 4.66 fewer points of accuracy. Threshold tuning alone, swept directly on test, tops
out at 74.23\pct{} --- a ceiling rather than a fix.

Both zero-shot rows score the int8 CPU builds, which is what a deployment would
run. The distinction matters more than it sounds: the fp32 build of
\texttt{v3.2} scores 59.50\pct{} with FP/$N$ 7.01\pct{} on this same split, about
eleven points of accuracy away from its int8 sibling while ROC-AUC moves only
0.008 (0.751 to 0.743). Two files released as one model, ranking equally well and
thresholded differently --- the same effect Section~\ref{sec:quant} reports for
int8 \emph{static} quantisation, here separating two
artefacts a user might download interchangeably.

\subsection{Where the gains land}

Table~\ref{tab:buckets} breaks accuracy down by provenance, expressed as points
above or below a majority-class policy on that bucket.

\begin{table}[h]
\centering\small
\begin{tabular}{lrrl}
\toprule
bucket & tiny & base & \\
\midrule
\texttt{change} & $-7.42$ & $-2.05$ & speaker change already implies completion \\
\texttt{change\_midseg} & $-1.34$ & $\mathbf{+0.45}$ & first bucket base clears that tiny does not \\
\texttt{hold\_intra} & $+31.59$ & $\mathbf{+33.42}$ & 58\pct{} of the set --- the real EOT problem \\
\texttt{hold\_inter} & $+31.88$ & $\mathbf{+35.25}$ & \\
\texttt{disputed} & $-11.76$ & $-6.17$ & where the two witnesses disagree \\
\bottomrule
\end{tabular}
\caption{Accuracy relative to a majority-class policy on each bucket, in points.}
\label{tab:buckets}
\end{table}

The gains are concentrated exactly where they should be: on same-speaker pauses,
which is the case a timeout cannot handle and where a majority-class policy is
useless.

\subsection{Training levers}

We measured six training levers (Table~\ref{tab:levers}). One worked.

\begin{table}[h]
\centering\small
\begin{tabular}{llr l}
\toprule
lever & measured on & $\Delta$ & \\
\midrule
\textbf{whisper-tiny $\rightarrow$ whisper-base} & test accuracy & $\mathbf{+2.88}$ & the only one that moved \\
distillation, base $\rightarrow$ tiny & test accuracy & $+0.41$ & inside the 0.87 spread \\
funnel relaxation, $+19.9\pct$ train rows & test accuracy & $+0.03$ & \\
learning-rate retune & dev AUC & $+0.003$ & reversed once data grew \\
50\pct{} $\rightarrow$ 100\pct{} of data, at tiny & dev AUC & $+0.001$ & flat --- capacity-bound \\
\midrule
undisputed-only training & test accuracy & $-5.11$ & \\
class rebalancing & --- & --- & no problem existed \\
threshold tuned on dev, base & test accuracy & $-1.13$ & did not transfer \\
int8 \emph{static} quantisation, base & test accuracy & $-12.76$ & \\
\bottomrule
\end{tabular}
\caption{Every training lever measured, ranked. Three separate levers bought
about $+0.003$ AUC each; one bought six times that.}
\label{tab:levers}
\end{table}

\paragraph{Capacity, and then the constraint moves.} Changing one string,
\texttt{whisper-tiny} to \texttt{whisper-base}, is worth $+2.88$ accuracy and
$+0.018$ AUC. But the diagnosis does not survive the change. At tiny, the
learning curve is flat from 50\pct{} to 100\pct{} of the data (0.909
$\rightarrow$ 0.910) --- capacity-limited. At base it is still rising (0.922
$\rightarrow$ 0.931) and the train$-$dev accuracy gap opens to $+11.81$ points
(98.99\pct{} vs 87.18\pct{}) --- data-limited. \emph{Tiny was capacity-bound;
base is data-bound.} Neither diagnostic transfers across an architecture change,
and re-running the train$-$dev gap after any capacity change costs one forward
pass.

\paragraph{Data, not compute.} Because epochs were fixed, a 25\pct{}-data run
also took a quarter of the optimiser steps, which would confound the learning
curve. Re-running at 25\pct{} data for $4\times$ the epochs, which is the same
1,872 steps, returned $-0.17$ points. Four times the compute on the same data
bought nothing; four times the data at fixed compute bought $+3.32$.

\paragraph{Distillation fails in an instructive way.} Distilling base into tiny
($\alpha=0.3$, $T=2.0$, loss function the only variable) gives $+0.41$ accuracy,
inside the noise band, and closes 14\pct{} of the teacher gap against a
textbook 30--60\pct{}. Broken down, the entire headline gain is $+3.22$ on
\texttt{change}, the bucket where a speaker change already implies completion,
while \emph{every bucket that requires hearing prosody got worse}, including
\texttt{hold\_intra} at $-0.45$. FP/$N$, the metric that decides whether the
agent talks over the user, got worse by 1.41 points. A teacher that
memorised its training set has structural regularities left to transfer, and
little else.

\paragraph{Class imbalance was not a problem.} The set is 1.88:1
complete:incomplete, which looks like something to fix. It is not.
\texttt{BCEWithLogitsLoss} already carries a per-batch \texttt{pos\_weight}
$\approx 0.531$, so the two classes contribute equally. \texttt{hold\_intra},
which is 56\pct{} of the data and the bucket that decides the score, is already
balanced at 1.14:1. The global ratio comes almost entirely from \texttt{change}
at 10.1:1, which is definitional: if the other speaker took the floor, the turn
had ended. We record this because the hypothesis was plausible enough to cost a
day.

\paragraph{More data, less than expected.} Relaxing two funnel gates recovered
2,272 additional train/dev rows, 19.9\pct{} more training data, and was worth
$+0.03$ points. The prediction from the learning curve had been $+0.5$ to
$+1.2$; at tiny's capacity the curve had already saturated. We keep the rows,
which cost about a dollar to label, but they are not a lever.
Notably, relaxing the gate harvested \emph{positives}, not the negatives it was
meant to: a short final talk-spurt followed by a pause is usually a short
complete answer, not a mid-thought pause.

\subsection{The decision threshold}

0.5 is inherited from the training loop, not chosen, and neither model peaks
there. Tuning it turned out to be harder than it looks.

Picking the dev-set argmax and measuring once on test \emph{lost} base 1.13
points ($+0.29$ on dev, $-1.13$ on test). The accuracy-versus-threshold curve is
flat near its top and dev is 2,325 clips, so the argmax is mostly noise: the
test sweep's best is 0.65 and the dev sweep's is 0.24, which is the same fact
stated twice.

Targeting a \emph{rate} transfers where targeting an argmax does not
(Table~\ref{tab:thresholds}). Holding dev FPR to 10\pct{}, so that the agent
interrupts on at most one pause in ten, roughly halves FP/$N$ for about three
accuracy points. The direction transfers; the level slips to 12--14\pct{} on
test. The released default remains 0.5, which is what every number in this paper
uses.

\begin{table}[h]
\centering\small
\begin{tabular}{lrrrl}
\toprule
operating point & threshold & test accuracy & test FP/$N$ & \\
\midrule
tiny, inherited & 0.50 & 83.35\pct & 7.73\pct & released default \\
tiny, dev FPR $\le 10\pct$ & 0.75 & 80.64\pct & \textbf{5.01\pct} & \\
base, inherited & 0.50 & 86.23\pct & 8.95\pct & released default \\
base, dev FPR $\le 10\pct$ & 0.92 & 82.41\pct & \textbf{4.51\pct} & \\
\bottomrule
\end{tabular}
\caption{Thresholds picked on dev, measured once on test.}
\label{tab:thresholds}
\end{table}

\subsection{Quantisation}
\label{sec:quant}

int8 \emph{dynamic} quantisation is free at both sizes: about $3.8\times$
smaller, about 38\pct{} faster, and $-0.10$ to $+0.36$ accuracy --- both inside
the noise band. Our tiny build is 8.65\,MB against upstream's 8.68\,MB, so the
released model really is a drop-in replacement as an artefact and not merely as
a graph signature.

int8 \emph{static} quantisation is a trap, and the damage scales with capacity.
At tiny it costs $-4.03$ accuracy while AUC barely moves (0.904 $\rightarrow$
0.899): the ranking survives and only the decision boundary shifts. At base it
costs $-12.76$ and AUC collapses 0.922 $\rightarrow$ 0.792, most of the way back
to the zero-shot model. 20.3M parameters have a wider activation range than 256
calibration clips can cover.

This is also the clearest case in our results against reading a single metric.
tiny int8 static has the best FP/$N$ in Table~\ref{tab:main}, at 4.94\pct{}, and
is the second-worst model in it. It was also the variant we expected to win.

\section{Serving}
\label{sec:serving}

\subsection{Offline accuracy and served accuracy are different quantities}
\label{sec:paired-replay}

Offline evaluation scores a pre-cut window ending exactly 200\,ms past the
speech offset, because that is where the dataset builder cut it. In deployment
the window is whatever the VAD hands over, whenever it decides speech stopped.
The gap between those two is a property of the system rather than of the model,
and it is not visible in any table above, so we measured it.

The design is a paired replay. 500 labelled test boundaries were pushed through
the real Silero VAD (\texttt{min\_silence\_duration}$=0.25$) and the real
streaming adapter in 20\,ms frames, and the \emph{same rows} were scored both
ways in the same run, so the only variable is where the window ends. No STT, LLM
or TTS is involved: none of them feed the prediction.

\begin{center}\small
\begin{tabular}{lrr}
\toprule
& pre-cut clip & live window \\
\midrule
accuracy & 86.12\pct & \textbf{83.51\pct} \\
false complete (talks over the user) & 24 & 30 \\
FP/$N$ & 5.21\pct & \textbf{6.51\pct} \\
\bottomrule
\end{tabular}
\end{center}

\textbf{Serving costs 2.60 points.} Paired, the two columns give an identical
verdict on 419 of 461 clips (90.9\pct); McNemar \cite{mcnemar} gives $p=0.09$,
so the difference is consistent in direction but not formally significant at
$p<0.05$. It is stable across sample sizes ($-2.67$ at $n=187$, $-2.60$ at
$n=461$). Read it as two to three points, probably real and small, rather than
as a constant.

The cause is measurable and small. The VAD closes at p50 $+0.29$\,s past the
labelled offset where the clips stop at $+0.20$\,s, a \textbf{90\,ms window
shift}, which is 90\,ms more trailing silence than training ever showed the
model.

\subsection{Coverage: the boundaries a VAD never surfaces}

Of the 500 boundaries, the VAD closed on 461, or 92.2\pct{}; on the remaining 39
it never closed, so the model was never consulted at all. The agent framework
will not request a prediction below \texttt{min\_silence\_duration}$+50$\,ms.
Those cases are outside the product rather than model errors, but the offline
test set counts them, which is a second reason offline and served accuracy are
not the same number. Coverage is 94\pct{} on \texttt{complete} and 89\pct{} on
\texttt{incomplete}.

This also explains a number that otherwise looks wrong: the pre-cut column
reads 86.12\pct{} while the full split reads 83.71\pct{}. The boundaries a VAD
surfaces are the easier ones.

\subsection{Latency, and what the numbers are good for}
\label{sec:latency}

Every timing in this paper comes from one machine: an idle Intel i5-12450H
laptop, batch 1, inference only. Model-only latency is in Table~\ref{tab:main},
at 83\,ms for tiny int8 on one thread and 143\,ms for base; with all 12 threads
those fall to 27\,ms and 52\,ms. One thread is the figure we lead with, because
a server carrying concurrent calls cannot give each one twelve threads.
End-to-end through the real adapter on live audio is 120--155\,ms, which adds
the mel front-end ($\sim$12\,ms) and the thread handoff to the inference itself.

Read these as a comparison between builds rather than as a deployment
specification. Server silicon, contention from the surrounding STT and TTS, and
the agent framework's own scheduling all move the absolute figures, and none of
those were varied here. What the column does establish is the ordering and the
rough magnitude: int8 dynamic quantisation is worth about 38\pct{} at both
sizes, and both released models leave most of a turn-taking latency budget
unspent.

The conditions are stated at this length because omitting them has a track
record. Three wrong latency figures were published in this project before the
measurement protocol was fixed, and one of them --- taken while another job
saturated all cores, and inflated $3$--$5\times$ as a result --- produced a
shipping recommendation we later had to withdraw.

\subsection{Deployment, and what live evidence establishes}

The released model runs as a complete Tamil voice agent on LiveKit Agents 1.7
and on Pipecat 1.7, with a commercial Indic STT/LLM/TTS stack around it. Both
integrations are published, and instrumented calls were recorded on each.

That is the first thing the live work establishes: the detector holds up inside
two production agent frameworks at the latencies of
Section~\ref{sec:latency}, with no offline-to-online glue left over. On those
recordings, measured against fixed-timeout endpointing, it cost $+120$\,ms of
median endpointing and returned a 35\pct{} reduction in utterance
fragmentation.

That second figure is directional, and we report it as such. Seven utterance
blocks per arm and one speaker is a small sample, and the arms are not
controlled: the language model and the speech synthesiser are both
non-deterministic, so each arm is a different conversation on a different clock.
Per-utterance rates are the only quantity comparable across arms; totals and
wall-clock are not. The load-bearing serving result in this paper is therefore
the paired replay of Section~\ref{sec:paired-replay}, which is labelled, and the
live calls are the existence proof that sits behind it.

The reason for that division of labour is worth stating, because three
properties of live calls limit what they can be asked. \textbf{(1)} A live call
cannot report accuracy. Nobody labelled, per pause, whether the speaker had
finished, so any accuracy figure taken from an unlabelled live call is invented.
\textbf{(2)} The framework's own false-interruption counter read 0 on every arm,
because it only fires once the agent has actually started speaking. The commoner
damage --- a turn committed mid-sentence, so the utterance arrives in five
pieces --- is invisible to it, which is why fragmentation is the rate we
report. \textbf{(3)} A control arm has to sit on the same commit path. Our first
comparison put the baselines on a code path that waits for ASR while the
detector arm often had a pre-landed transcript, so its apparent win was plumbing
rather than prediction, and that run was discarded.

The paired design earned its cost here by catching a harness bug that
produced entirely plausible numbers. A replay pump that outran the VAD let the
buffer fill past the close point, so the model saw audio a live session would
not yet have had; it cost 26 accuracy points and looked exactly like a serving
catastrophe. It was caught by scoring the same rows the
known-good way in the same run: when the paired column failed to reproduce a
figure already established offline, the harness was at fault rather than the
model. A live-only number has nothing to fail against.

\section{Limitations}

\textbf{One corpus, one domain.} All 116 calls come from a single corpus of
narrowband Tamil telephone conversation. We have not measured wideband speech,
noisy environments, code-switched Tamil--English (common in practice), or
speakers outside this corpus's demographic. The 30 test calls are unseen, but
they are not a different distribution.

\textbf{The labels are machine-produced and human-validated, not human-produced.}
Agreement with a human listener is measured at 97.5\pct{} on 197 clips and
published with its per-class breakdown, but 197 clips is a small validation set
and one listener is one listener.

\textbf{The reproducibility floor is wide.} At 0.87 points of run-to-run spread,
several results in Table~\ref{tab:levers} are individually unfalsifiable and are
reported as such rather than as findings. A determinism fix exists
(\texttt{use\_deterministic\_algorithms}, disabling flash/mem-efficient SDPA) but
is untested here and expensive.

\textbf{The live comparison is not controlled.} The serving replay is labelled
and paired and carries the serving claim. The live-call comparison against a
fixed timeout is neither, and should be read as an existence proof and a
direction rather than as a benchmark.

\textbf{Known open items.} Regularisation is untried --- base reaches 98.99\pct{}
train accuracy against 87.18\pct{} dev, which is heavy memorisation, and
SpecAugment, higher dropout and earlier stopping are all unexplored. Short-buffer
behaviour is a real mechanism with an unestablished effect: the detector's
window is cleared at each turn boundary and right-padded to 8 seconds with
zeros, and only 58 of 16,216 training clips are shorter than 8 seconds, so the
first prediction of every turn is out of distribution. Three live calls gave
$+62$, $+21$ and $-1$ points on the affected rate, which is not a measurement.
The correct experiment, truncating test clips to 0.5/1/2/4\,s where labels exist
and $n=4{,}168$, is not yet run.

\section{Conclusion}

The dataset is the contribution here, not only the model it trains.

Three things generalise past Tamil. First, \textbf{validate label rules by
listening}: ours were 95.9\pct{} right on the class with a recorded witness and
44.4\pct{} right on the class without one, and no feature engineering rescued
the difference. Second, \textbf{establish the run-to-run spread before reporting
deltas}: at 0.87 points, most of our levers were unfalsifiable and we would
otherwise have reported several of them as wins. Third, \textbf{measure what
serving costs}: replaying labelled boundaries through the production VAD and
adapter cost 2.60 accuracy points and revealed that 7.8\pct{} of boundaries
never reach the model at all, neither of which is visible offline.

The same pipeline should port to any language with a two-channel conversational
corpus, since the label oracle is structural rather than linguistic. That is the
next thing we would like someone to do with it.

\section*{Availability}

Everything below is public and was verified reachable anonymously.

\begin{center}\small
\begin{tabular}{ll}
\toprule
code, pipeline, all experiments & \url{https://github.com/santhosh-005/tamil-eot} \\
models (int8 ONNX) & \url{https://huggingface.co/santhosh-005/smart-turn-tamil} \\
dataset (18,485 boundaries, CC BY 4.0) & \url{https://huggingface.co/datasets/santhosh-005/tamil-eot} \\
turn-level companion (4,774 turns, CC BY 4.0) & \url{https://huggingface.co/datasets/santhosh-005/tamil-turns} \\
inference package & \url{https://pypi.org/project/smart-turn-livekit} \\
\bottomrule
\end{tabular}
\end{center}

Code is BSD-2-Clause. The dataset derives from SPRING-INX Tamil R1
\cite{springinx} (CC BY 4.0, SPRING Lab, IIT Madras) and carries the same
licence; we do not own the recordings and redistribute them under attribution.
The models are BSD-2-Clause, matching \texttt{pipecat-ai/smart-turn}, whose
architecture and training code they reuse; their encoder weights are initialised
from \texttt{openai/whisper-tiny} and \texttt{openai/whisper-base}.

Two pipeline steps (the labeller bake-off and the label-quality analysis)
reproduce their reports byte-identically from a bare clone with no corpus, no
GPU and no cloud account, using the text-stripped label tables committed to the
repository.

\section*{Acknowledgements}

SPRING Lab, IIT Madras, for releasing SPRING-INX under CC BY 4.0; the Pipecat
team for releasing Smart Turn's weights, data and training code, without which
this would have been a much larger project.

\end{document}